\documentclass[10pt,journal,compsoc]{IEEEtran}

\usepackage{graphicx}
\usepackage{amssymb}
\usepackage{amsmath}
\usepackage{latexsym}
\usepackage{multirow}
\usepackage{url}
\usepackage{hyperref}
\usepackage{pifont} 
\ifCLASSOPTIONcompsoc
  \usepackage[nocompress]{cite}
\else
  \usepackage{cite}
\fi

\begin{document}
\title{Combining Deep Learning Classifiers\\for 3D Action Recognition}
%
\author{Jan Sedmidubsky and Pavel Zezula
\IEEEcompsocitemizethanks{
\IEEEcompsocthanksitem Affiliation: Masaryk University, Botanicka 68a, 602\,00 Brno, Czechia
\protect\\
Email: xsedmid@fi.muni.cz
}
\thanks{Manuscript submitted: April 21, 2020}
}

\markboth{Submitted to Pattern Recognition Letters}%
{Sedmidubsky \MakeLowercase{\textit{et al.}}}

\IEEEtitleabstractindextext{%
\begin{abstract}
The popular task of 3D human action recognition is almost exclusively solved by training deep-learning classifiers. To achieve a high recognition accuracy, the input 3D actions are often pre-processed by various normalization or augmentation techniques. However, it is not computationally feasible to train a classifier for each possible variant of training data in order to select the best-performing subset of pre-processing techniques for a given dataset. In this paper, we propose to train an independent classifier for each available pre-processing technique and fuse the classification results based on a strict majority vote rule. Together with a proposed evaluation procedure, we can very efficiently determine the best combination of normalization and augmentation techniques for a specific dataset. For the best-performing combination, we can retrospectively apply the normalized/augmented variants of input data to train only a single classifier. This also allows us to decide whether it is better to train a single model, or rather a set of independent classifiers.
\end{abstract}

}

\maketitle

\IEEEdisplaynontitleabstractindextext

\IEEEpeerreviewmaketitle


\section{Introduction and Related Work}

A human motion can be described by a sequence of skeleton poses, where each pose keeps 3D coordinates of important body joints in a specific time moment. Such spatio-temporal data can be acquired using dedicated hardware technologies or recent pose-estimation methods capable of determining 2D or even 3D joint positions from ordinary videos~\cite{SXLW19}. The acquired data have a great potential to be employed in many application fields, e.g., in sports to automatically detect fouls during a football game;
in security to detect potential threats like a running group of people; or in computer animation to recognize previously-captured animations.

\emph{Action recognition}
is the most popular motion-processing task that aims at determining the class of pre-segmented actions based on a labelled set of training ones. Solving this task is challenging as the actions of the same class can be performed by different subjects in various styles, speeds, and initial body postures.
The variability of actions is often decreased by applying various \emph{normalization} techniques~\cite{WW17,PZHT14}.
The normalized actions are then used to train a deep neural network classifier.
To make deep learning more robust, the training data are enriched using \emph{augmentation} techniques~\cite{WCSPLB19,KBASB17}. However, the question is how to select a suitable combination of pre-processing techniques for a given dataset.


\subsection*{Related Work}

Traditional action-recognition methods based on handcrafted features have a limited ability to represent the complexity of spatio-temporal movement patterns and have been practically forgotten due to the progress in deep learning~\cite{WCHPH19,KBRASB20}. In deep learning, the input actions are usually transformed into intermediate representations (e.g., graph structures~\cite{ZWSJ19,LXZZCY19}, 2D motion images~\cite{LBTD17,SEZ18-mtap}, or histograms~\cite{ACHCS18}) that are then used to train a classifier, based on convolutional neural networks (CNN)~\cite{AK18,LBTD17}, graph convolutional networks (GCN)~\cite{LGKQG19,LXZZCY19}, or Long Short-Term Memory (LSTM) networks~\cite{WWD19,LWDHK18}.

To enhance the recognition accuracy, different classifiers or data modalities can be \emph{fused}. The fusion approach in~\cite{NCPMV18} proposes to learn features of individual 3D skeletons using CNN and then train a LSTM network on top of a sequence of such features. In~\cite{WWD19}, multi-modal features are firstly extracted from the input actions and then fused by an autoencoder network. In~\cite{LQYZS20}, the authors even propose to fuse the RGB and 3D skeleton modalities.


To sum it up, there exist over a hundred papers that propose complex neural-network classifiers and various data normalization or augmentation techniques. This implies it is computationally unrealistic to train a model for every variant of input actions generated by a specific combination of pre-processing techniques (even for a fixed neural-network architecture).
Moreover, the success of each combination of pre-processing techniques depends on a given dataset.


\subsection*{Contributions of this Paper}


Our objective is to determine the best combination of data pre-processing techniques for a given classifier and dataset.
The main idea lies in training only a single independent model for each pre-processing technique and applying the fusion approach to estimate the quality of a selected combination of techniques, instead of training several orders of magnitude more classifiers.
%
Specifically, we introduce these contributions:
\begin{itemize}
    \item We propose two effective augmentation techniques for 3D actions (called BodyModel and KeyPose augmentations);
    \item We design an online fusion of classifiers based on a strict majority vote rule;
    \item We introduce an algorithmic procedure for efficient evaluation of a very large number of combinations of classifiers.
\end{itemize}


In addition, for the best combination of pre-processing techniques, we can retrospectively apply the normalized and augmented variants of input data to train only a single \emph{all-in-one} model. We expect that such model can be confused by many variants of training data, which can lead to a decreased recognition accuracy in comparison with the fusion of independent classifiers that are trained for the specific purpose.



\section{Data Pre-Processing Techniques}
\label{sec:dataPre-processing}

We represent skeleton data of a single \emph{action} as a sequence $(P_1, \ldots, P_l)$ of $l$ consecutive 3D \emph{poses} $P_i$, where the $i$-th pose $P_i \in \mathbb{R}^{j \cdot 3}$ is captured in time moment $i$ ($1 \le i \le l$) and consists of $xyz$-coordinates of $j$ tracked \emph{joints}. In this paper, we use three variants of body models with $j \in \{12, 14, 31\}$ joints.
To improve the quality of deep learning, we apply the following normalization and augmentation techniques.


\subsection{Action Normalizations}
\label{sec:normalization}

The semantically equivalent actions, i.e., belonging to the same class, can be performed by subjects (i) at different space locations, (ii) facing various directions, or (iii) having different heights. Since absolute positionings and orientations and different subject sizes rather introduce unwanted bias for recognizing daily/exercising actions~\cite{WW17,PZHT14}, we use the following normalization techniques.
%
\begin{itemize}
    \item {\bf P}-normalization -- To unify actions performed at different space locations, each skeleton pose $P_i$ is shifted into a skeleton-centric coordinate system so that the root joint $P^{root}_i$ is aligned to the origin $(0,0,0)$.
    \item {\bf O}-normalization -- To unify various subject orientations, each pose is rotated to align the line connecting both left and right hip joints to be parallel with the $x$-axis.
    \item {\bf S}-normalization -- Subjects are unified in sizes so that each skeleton is resized to the height of an ``average'' human.
\end{itemize}

All the three normalizations help to reduce the spatial variability in joint coordinates over the actions in the same class, and thus facilitate the neural-network training process.


\subsection{Action Augmentations}
\label{sec:augmentation}

\begin{figure}
  \includegraphics[width=\columnwidth]{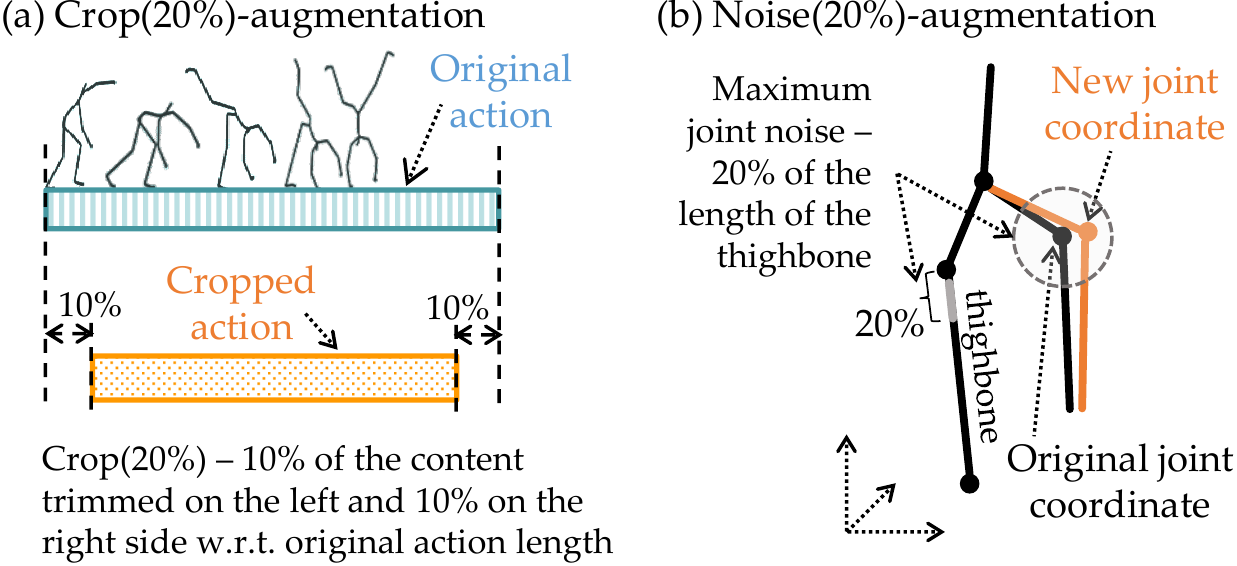}
  \caption{Crop- and Noise-augmentation techniques: (a) cropping the original action by $20\,\%$ and (b) moving a joint into a new random position, which is at most $20\,\%$ of the thighbone length away from the original position.}
  \label{fig:augmentation-Crop_Noise}
\end{figure}

\begin{figure}
  \includegraphics[width=\columnwidth]{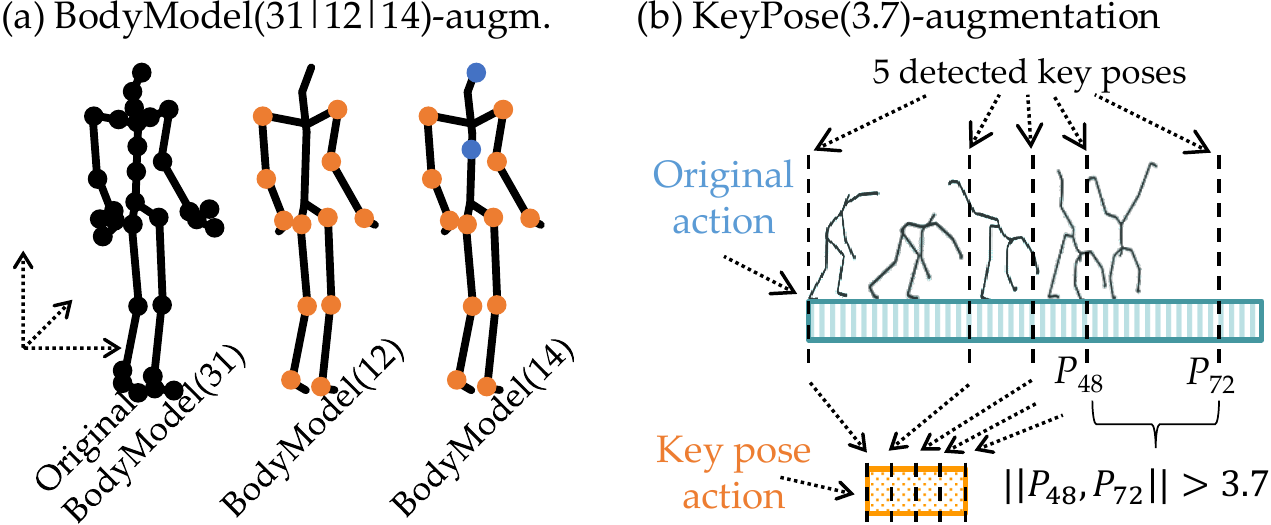}
  \caption{Proposed BodyModel- and KeyPose-augmentation techniques: (a) simplifying the body model from the original 31 joints to selected 12 and 14 joints and (b) considering only key poses of the original action.
  }
  \label{fig:augmentation-BodyModel_KeyPose}
\end{figure}

Existing deep learning classifiers achieve a high recognition accuracy if a sufficiently large number of training actions is provided. However, action datasets might contain only a limited number of samples in each class, e.g., as in the HDM05 dataset~\cite{MRCEKW07} providing
less than 20 actions per class. To enlarge training data,
we utilize the following two augmentation techniques, originally proposed by~\cite{SZ19-ism}.
\begin{itemize}
    \item {\bf Crop}($range$)-augmentation -- Each action is cropped by trimming away its left and right side. The $range$ parameter (in percents) determines that $\frac{range}{2} \cdot l$ poses are cut from the left side and the same amount from the right side, with respect to the action length $l$. This technique keeps the most important middle part of actions, while the boundary parts that need not be well segmented are discarded.
    %
    \item {\bf Noise}($range$)-augmentation -- A noise is added into 3D coordinates of all action joints simply by moving the joint position in each of $x$/$y$/$z$ axis by a random value. The $range$ parameter (in percents) bounds the maximum size of the random value with respect to the average length of thighbone, i.e., to the maximum $range \cdot length_{thighbone}$.
    %
\end{itemize}
Both the Crop- and Noise-augmentation techniques are graphically illustrated in Fig.~\ref{fig:augmentation-Crop_Noise}.
In addition, we propose the following two new augmentation techniques, illustrated in Fig.~\ref{fig:augmentation-BodyModel_KeyPose}.
\begin{itemize}
    \item {\bf BodyModel}($model$)-augmentation -- The body model with 31 joints is simplified by selecting only a subset of joints, specified within the $model$ parameter (see Fig.~\ref{fig:augmentation-BodyModel_KeyPose}a).
    This should facilitate learning since the spatial complexity of each pose is reduced by ignoring some of very close joints that produce an unnecessary movement noise.
    \item {\bf KeyPose}($dist$)-augmentation -- The original frame-per-second (FPS) rate is non-linearly decreased by considering only specific poses, so-called \emph{key poses}. The first action pose is always considered as the key pose and the other ones are gradually determined as the closest next pose which is sufficiently dissimilar, i.e., the dissimilarity between the current and previous key pose is higher than the $dist$ parameter. The dissimilarity of two poses is quantified as the sum of Euclidean distances between their corresponding pairs of 3D joint coordinates.
    Compared to traditional downsampling techniques with fixed FPS rates, this technique better respects the changes in movement.
\end{itemize}

While the Crop- and KeyPose-augmentation techniques deform the temporal dimension of original actions, the Crop- and BodyModel-augmentations change their spatial domain. Except for the Noise-augmentation, the other remaining three techniques can significantly reduce the original size of actions.
Since each of the four augmentation techniques is parameterizable, we can generate several variants of actions using only a single technique, e.g., by applying Crop($10\,\%$) and Crop($20\,\%$).


\section{Action Recognition using a Single Bi-LSTM Classifier}
\label{section:lstm}

The normalization and augmentation techniques can be applied to generate a set of training actions for learning a classifier. As a classifier, we adopt a light-weight version of \emph{bidirectional Long-Short Term Memory} (Bi-LSTM) network. Formally, we classify input action $(P_1, \ldots, P_l)$ into one of $m$ predefined classes $\{C_1, \ldots, C_{m}\}$, where each class $C_c \, (c \in [1, m])$ is characterized by a non-empty set of training actions.
%
We first embed each 3D action pose $P_i \in \mathbb{R}^{j \cdot 3}$ into an $E$-dimensional space by a linear projection (with parameters $W_E \in \mathbb{R}^{j \cdot 3 \times E}$ and $b_E \in \mathbb{R}^{E}$) followed by a ReLU activation:
\begin{equation}
\label{eq:embed}
P'_i = \textrm{ReLU}(W_E \cdot P_i + b_E),
\end{equation}
where $E \in \mathbb{N}$ is a user-defined parameter. Learning a projection of the original data in the end-to-end training phase permits us to work with lower-dimensional data and a higher level of abstraction,
with both effectiveness and efficiency advantages with respect to the original skeletons.

Each embedded pose $P'_i \in \mathbb{R}^{E} \, (i \in [1,l])$ is then fed to both the past-to-future and future-to-past LSTM cells, which respectively produce the following hidden state vectors $h_i, h'_i \in \mathbb{R}^{H/2}$:
\begin{equation}
h_i = \textrm{LSTM}(P'_i, h_{i-1}) \qquad
h'_i = \textrm{LSTM}(P'_i, h'_{i+1}),
\end{equation}
where $H \in \mathbb{N}$ is a user-defined parameter denoting the total feature size, i.e., the sum of dimensions of both the state vectors. The initial states $h_0$ and $h'_{l+1}$ are set to zeros. The state vector $h_i$, together with the consecutive embedded pose $P'_{i+1}$ (and similarly $h'_i$ with $P'_{i-1}$ for the reverse direction), are given as input to the next step.

\begin{figure}[tb]
  \includegraphics[width=\columnwidth]{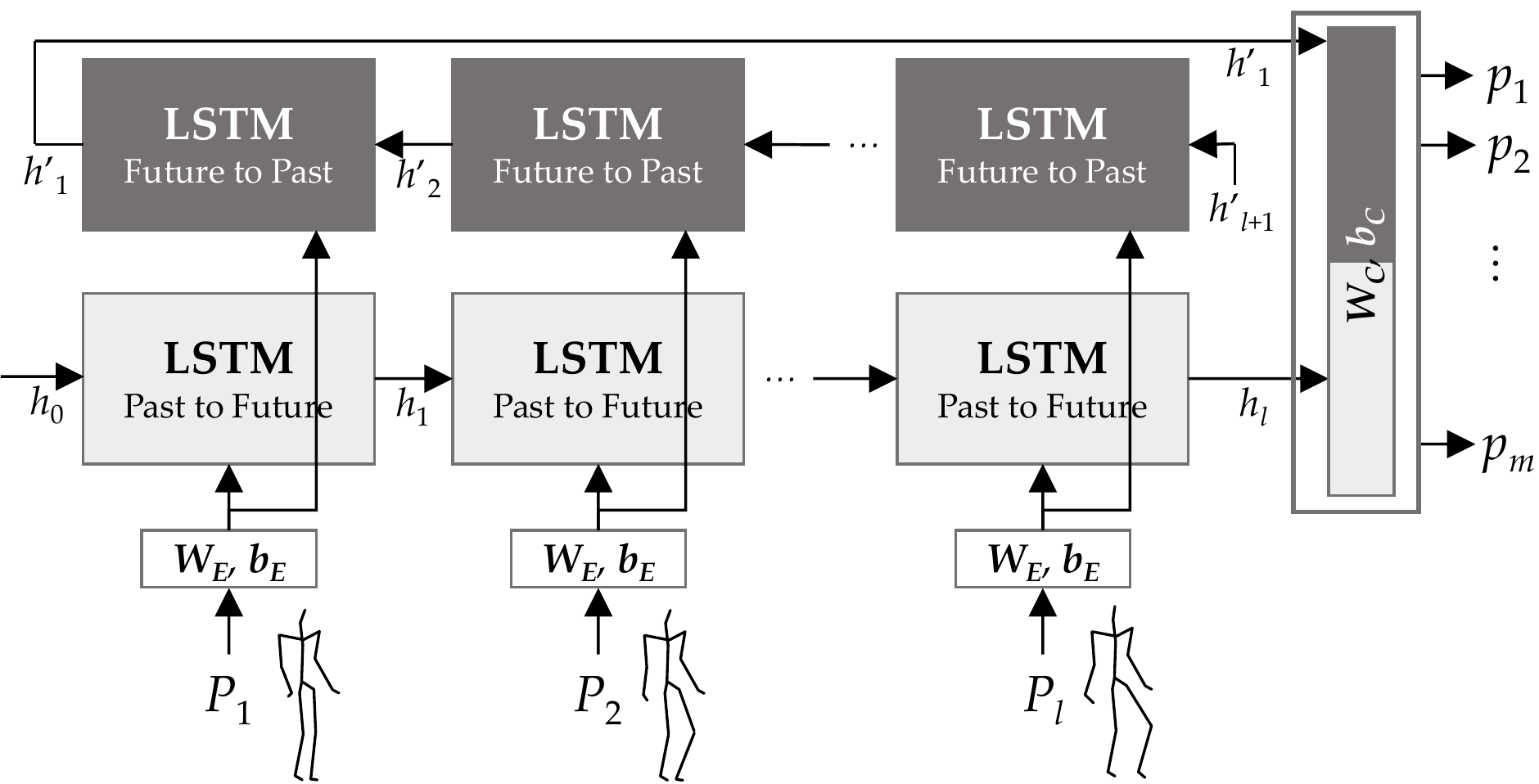} 
  \caption{Schema of a single Bi-LSTM classifier used for action recognition.}
  \label{fig:BiLSTM_model}
\end{figure}

For the given action, the prediction for each class $\mathcal{C}_c \, (c \in [1,m])$ is quantified by probability $p_c$ that is obtained from the concatenation $[h_l|h'_1]$ of hidden states $h_l$ and $h'_1$ as follows:
\begin{equation}
\label{eq:bi-prediction}
p_c = \sigma_c(W_\mathcal{C} \cdot [h_i|h'_i] + b_\mathcal{C}) \qquad c \in [1,m],
\end{equation}
where $W_\mathcal{C} \in \mathbb{R}^{H \times m}$ and $b_C \in \mathbb{R}^{m}$ are the parameters of a linear projection with $m$ outputs and $\sigma_c(\cdot)$ denotes the result of the softmax function applied to the $c$-th component of its argument. The class $C_c$ with the highest softmax value is considered as the classification result, i.e., $\forall i \in [1,m]: p_c \ge p_i$.
%
%
We optimize the parameters ($W_E$, $b_E$, $W_\mathcal{C}$, $b_\mathcal{C}$, and LSTM parameters) by minimizing the cross-entropy between the predictions and the targets.
The whole architecture is illustrated in Fig.~\ref{fig:BiLSTM_model}.


\section{Online Fusion of Independent Classifiers}
\label{sec:fusion}

When multiple normalization and augmentation techniques are considered, training a single classifier for a given dataset is a hard task because of an extremely large number of variants how training actions can be pre-processed.
Suppose $n$ techniques are available, then there are $2^n$ possible combinations of techniques that generate different sets of training data. For example, if $n = 16$ (e.g., the four presented augmentation techniques and each of them parameterized in four different variants), then there are $2^{16}=65,536$ different subsets of combinations. And it is not computationally feasible to train such number of Bi-LSTM classifiers for choosing the best combination of pre-processing techniques for each specific dataset.

\begin{figure*}
\centerline{\includegraphics[width=\textwidth]{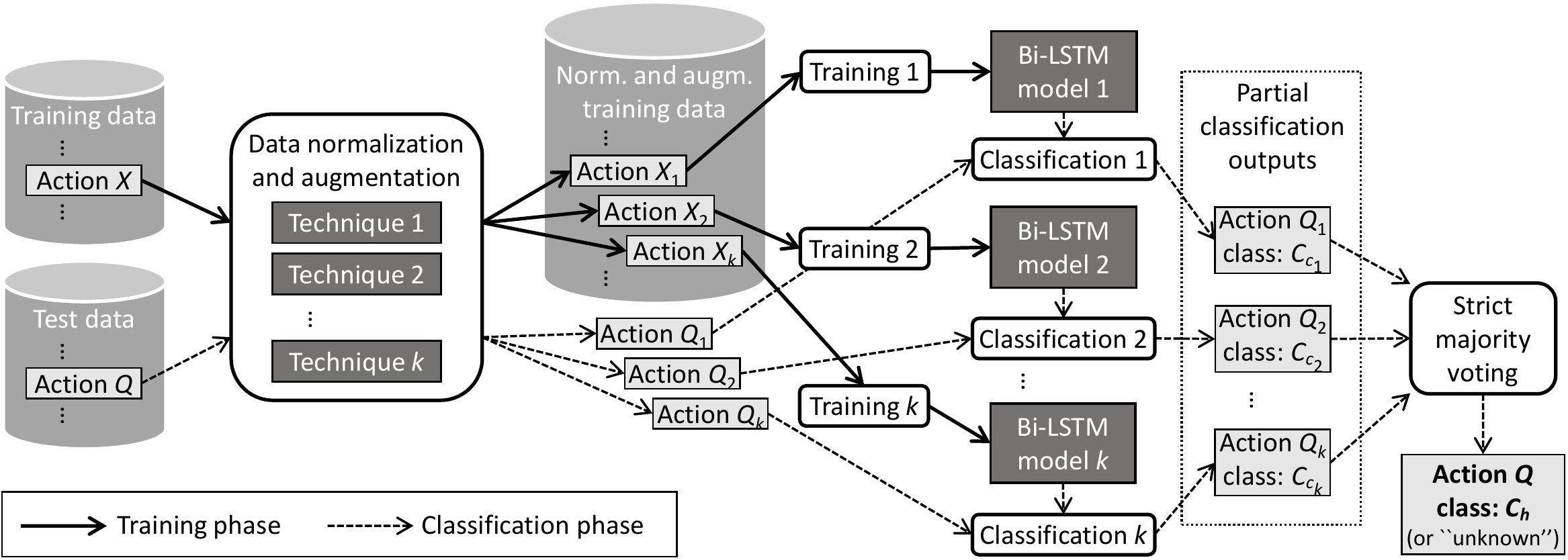}}
\caption{Schematic illustration of the fusion approach. In the training phase, each action $X$ is pre-processed by $k$ normalization/augmentation techniques to generate $k$ modified instances $X_1, \ldots, X_k$, that are further used to train $k$ independent Bi-LSTM classifiers. In the classification phase, each test action $Q$ is pre-processed in the same way and $Q_1, \ldots, Q_k$ modified instances are independently classified by the $k$ trained Bi-LSTM classifiers to get the partial outputs $C_{c_1}, \ldots, C_{c_k}$. Such outputs are processed by the strict majority vote rule to obtain the final classification $C_h$.}
\label{fig:schemaIndependentClassifiers}
\end{figure*}

Instead of training $2^n$ combinations, our idea is to train only $n$ independent Bi-LSTM classifiers, i.e., a single classifier for each pre-processing technique. Then, we efficiently estimate the quality of a specific combination of $k \in \mathbb{N} \, (k \le n)$ classifiers by evaluating the test-data accuracy using an \emph{online fusion} approach. In particular, each test action $Q$ is classified using the following three-stage process:
\begin{enumerate}
    \item The test action $Q$ is normalized/augmented by the given $k$ pre-processing techniques to get the modified $k$ action instances $Q_1, \ldots, Q_k$;
    \item The modified action instances are independently classified by the corresponding Bi-LSTM classifiers to get the $k$ partial classification outputs;
    \item The partial outputs are processed based on the majority vote principle to determine the final classification of $Q$.
\end{enumerate}
%
The whole process is schematically illustrated in Fig.~\ref{fig:schemaIndependentClassifiers}.


\subsection{Strict Majority Vote Principle}
\label{sec:majorityVote}

Traditionally, the class which receives the largest number of votes is selected as the consensus (majority) decision~\cite{KHDM98}. We use a much more strict version when more than half of the $k$ independent classifiers have to agree on the same class. If there is no class with $> \lfloor k/2 \rfloor$ votes, the classification result is considered as ``unknown''. Formally, we define the \emph{strict majority vote} principle for the output classes $C_{c_1}, \ldots, C_{c_k} \, (c_i \in [1,m])$ of $k$ independent classifiers as:
\begin{equation}
\left\{
\begin{array}{ll}
C_{h} & \quad \left|\,h \in \{c_1, \ldots, c_k\}\,\right| > \left\lfloor\frac{k}{2}\right\rfloor,\\
\text{``unknown''} & \quad \text{otherwise},
\end{array}
\right.
\label{eq:majorityVote}
\end{equation}
where $\{c_1, \ldots, c_k\}$ stands for the \emph{multiset} of indexes of partial output classes and $h$ determines the index of the $C_{h}$ class with the highest number of votes:
\begin{equation}
h = \underset{c_i \in [1,m]}{\operatorname{arg max}} \left|c_i \in \{c_1, \ldots, c_k\}\right|.
\nonumber
\end{equation}

Noticeably, the ``unknown'' output is always considered as misclassification when evaluating the test-data accuracy for the specific combination of classifiers. This implies that the used strict majority vote principle cannot exceed the accuracy of the traditional vote rule~\cite{KHDM98}. On the other hand, the result has a quite high confidence and can be very efficiently evaluated as described in the following.


\subsection{Evaluating All Combinations Efficiently}
\label{sec:efficientFusionEvaluation}


%
%
Having $n$ pre-processing techniques, we need to evaluate the accuracy of $2^n$ combinations. Naively, for the specific combination of $k$ techniques $(k \le n)$, each test action needs to be $k$-times classified by independent Bi-LSTM classifiers. This results in the following huge number of classifications:
\begin{equation}
\sum_{k=1}^{n} k \cdot \binom{n}{k} = n \cdot 2^{n-1}.
\end{equation}
Assume $n=16$ techniques and the batch of $1,164$ test actions (as later used in the experiments), we would need to perform about 610 million classifications. This would probably require roughly one month of a single graphics-card capacity.

However, each action is repeatedly processed by the same classifiers, which has to inherently lead to the same partial classifications. Therefore, we classify each action only once for each pre-processing technique, i.e., $n$-times in total, and store the partial classification results to disk. In this way, we can keep the list of partial outputs of all test actions for each out of $n$ pre-processing techniques. We can further filter these lists to store only the true-positive matches, i.e., keeping only the actions with the correct partial class assigned.

In the evaluation phase for the given combination of $k$ techniques, we simply merge the corresponding $k$ true-positive lists and count how many times each test action is present. Then, we filter out the actions having count equal or less than $\lfloor k/2 \rfloor$, which is the condition of the strict majority vote rule (see Equation~\ref{eq:majorityVote}). This allows us to very efficiently determine the class of each test action using the fusion approach. The test-data accuracy is finally determined as the ratio between the number of retained actions and the number of all test actions.

It is important to realize that this trick enables evaluating all combinations very efficiently. In particular, by storing the partial outputs, we save exactly $2^{n-1}$ classifications in contrast to the naive approach. In our case, the naive number of 610\,M classifications is reduced to 19\,K ($16 \cdot 1,164 = 19\,$K).


\section{Experimental Evaluation}

In the following, we present a test dataset and methodology for training Bi-LSTM classifiers. Then, we determine interesting parameters of the presented normalization/augmentation techniques and their combinations, which results in definition of 16 pre-processing techniques. Next, we train 16 corresponding classifiers, efficiently evaluate all their $2^{16}$ combinations using the fusion approach, and determine the most useful techniques based on the best-performing combinations. We finally apply the best techniques to train a single all-in-one model as an alternative to independent classifiers.


\subsection{Dataset}
\label{sec:dataset}

We use the popular HDM05 dataset~\cite{MRCEKW07} captured
at a 120 frame-per-second (FPS) rate. The dataset provides the ground truth that divides 2,345 actions into 130 classes.
For our internal experiments we ignore the 8 least populated classes that provide only from 1 to 4 action samples, which results in 2,328 actions in total.
The actions correspond to daily/exercising activities and significantly differ in length, ranging from 13 frames (0.1\,s) to 900 frames (7.5\,s).
%
We consider the HDM05 dataset as very challenging since it contains the highest number of 130 classes, while providing only less than 20 action samples for each class on average, in comparison with other datasets.
%


\subsection{Methodology}
\label{sec:methodology}

For a fair evaluation, we apply the standard 2-fold cross validation procedure within all the experiments. We split the
ground truth into two folds in a balanced way so that each fold contains 1,164 actions and roughly the same number of actions of the same class.
The experiment accuracy is always determined as the average over the best accuracies achieved in both fold runs.
%
%



We train Bi-LSTM models using 10-times down-sampled data of 12 FPS rate (except for the KeyPose augmentation with a variable FPS rate), since this rate is sufficient to retain main characteristics of actions, while reducing training time a lot.
The dimensionalities of the embeddings and of the combined hidden state vector are $48$ and $1,024$, as they achieve a reasonable trade-off between the accuracy and performance.
The training of a single model -- with $1,164$ training actions and $150$ epochs -- takes roughly $45\,$minutes, when performed on the NVIDIA Quadro K$1200$ graphics card.


\subsection{Evaluating the Best Combinations within 16 Classifiers}

To evaluate the fusion approach, we train $n=16$ independent Bi-LSTM classifiers. While the architecture of classifiers is still the same, they use differently pre-processed training/test data. In particular, we use 3 variants of normalizations: skeleton-centric (P), full (P+O+S), and without any normalization (--). Since the full normalization is expected to contribute to the best classification accuracy~\cite{PZHT14}, we apply augmentation techniques only to the P+O+S-normalized data. Specifically, we apply the four augmentation techniques introduced in Section~\ref{sec:augmentation}, each of them parameterized in two or three settings (e.g., settings KeyPose(10.6) and KeyPose(3.7) generate the actions of approximately 8 and 24\,FPS rate, respectively). All these settings correspond to 12 different pre-processing techniques and thus 12 classifiers are trained -- see the normalization and augmentation settings in rows 1--12 in Table~\ref{tab:combinationsOfClassifiers}.
Next two rows (13--14) denote classifiers with non-augmented training data, while test data cropped either by 10\,\%, or 20\,\%. The last two rows (15--16) correspond to the opposite variant when the cropped actions are used for training, while the non-augmented data for testing.

\begin{figure}
\centerline{\includegraphics[width=\columnwidth]{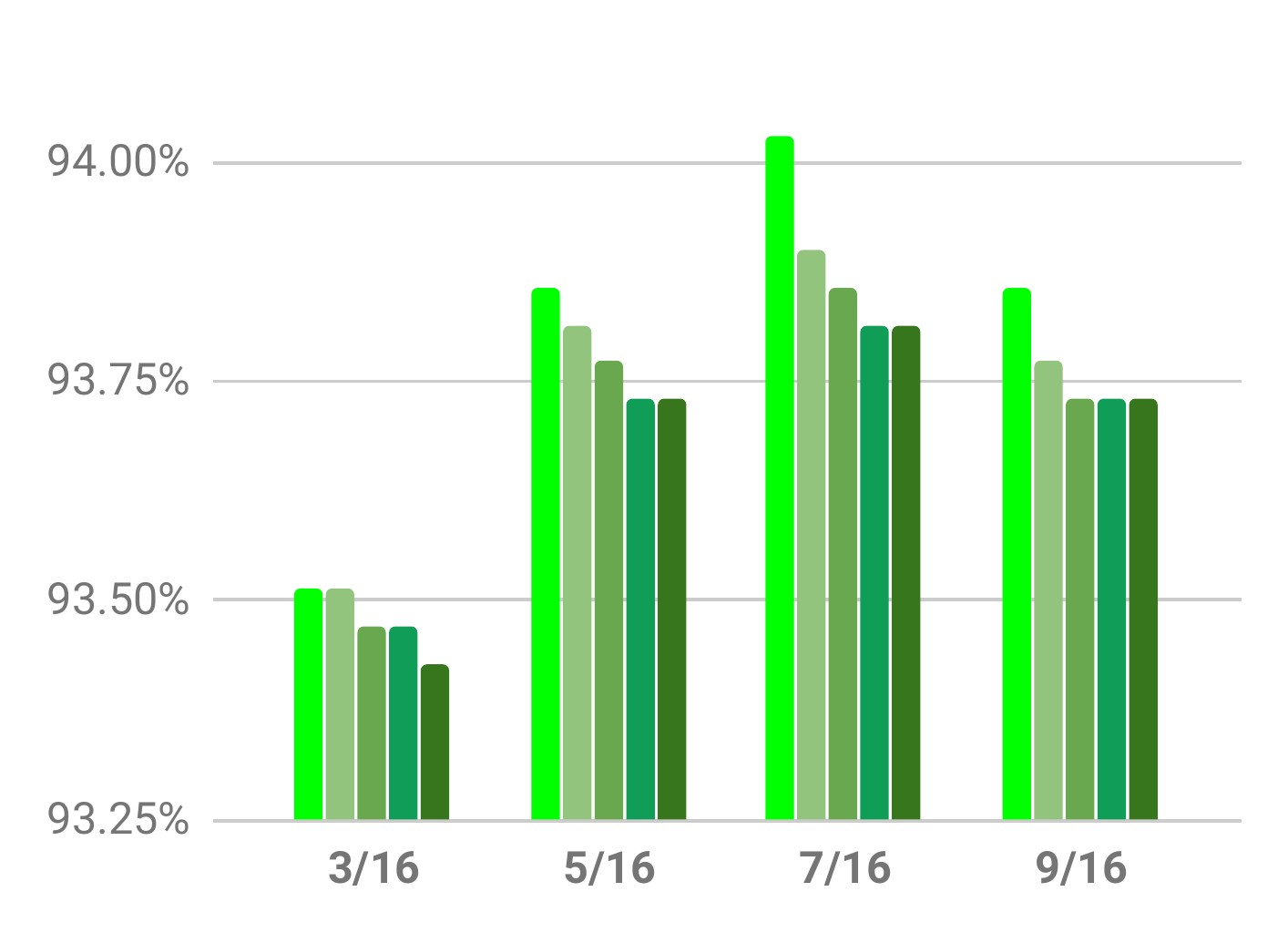}}
\caption{Accuracy of the fusion for 20 selected combinations (for each $k \in \{3, 5, 7, 9\}$, the 5 best combinations within $n=16$ classifiers are plotted).}
\label{fig:combinationsAccuracyChart}
\end{figure}

Firstly, we have trained such 16 standalone Bi-LSTM classifiers and evaluated their recognition accuracy -- see the fourth column (``Accuracy'') of Table~\ref{tab:combinationsOfClassifiers}. We can see that the best accuracy of 92.40\,\% is achieved when both training and test data are P+O+S-normalized and Noise(2.5\,\%)-augmented. Then, we have fused these 16 classifiers to efficiently evaluate all their $2^{16}$ combinations (see Section~\ref{sec:efficientFusionEvaluation} for more details) and illustrated the results of selected combinations in the right side of Table~\ref{tab:combinationsOfClassifiers}. In particular, we present the five combinations achieving the best results for $k \in \{3, 5, 7, 9\}$. For each selected combination, the involvement of the specific $k$ classifiers is denoted by black points and the final fusion accuracy is reported at the bottom of the table (for better clarity, the fusion accuracies are also graphically plotted in Fig.~\ref{fig:combinationsAccuracyChart}). We can see that all the reported fusion accuracies are above 93.40\,\%, with the maximum at 94.03\,\%. As expected, this experiment confirms superiority of the fusion approach that clearly outperforms any of the standalone classifiers, by reducing the recognition error by 21\,\% with respect to the best Noise(2.5\,\%)-classifier.

Interestingly, by focusing on black points in Table~\ref{tab:combinationsOfClassifiers}, we can observe that the non-normalized data and P+O+S-normalized BodyModel(14) and KeyPose(3.7) augmentations are included in most of the best-performing combinations, even if their corresponding standalone classifiers do not achieve convincing results. This demonstrates a big strength of the proposed approach that can automatically select the pre-processing techniques suitable for a given dataset.

\begin{table*}[t]
\setlength{\tabcolsep}{4px}
\caption{Comparison of action recognition accuracy of 16 independent classifiers (in rows) and their 20 selected combinations (in columns). For each out of 4 combination types (i.e., 3/16, 5/16, 7/16 and 9/16), the five best-performing combinations of classifiers are reported. The black points denote the involvement of independent classifiers that are used in the specific combination.
}
\label{tab:combinationsOfClassifiers}
\centering
\begin{tabular}{c|c|c|c|r||c|c|c|c|c||c|c|c|c|c||c|c|c|c|c||c|c|c|c|c|}
\hline
\multirow{2}{*}{\#} & \multirow{2}{*}{\bf Norm.} & \multicolumn{2}{c|}{\bf Augmentation of} & \multirow{2}{*}{\bf Accuracy} & \multicolumn{20}{c|}{\bf Combinations of independent classifiers} \\
& & {\bf training data} & {\bf test data} & & \multicolumn{5}{c||}{\bf 3/16} & \multicolumn{5}{c||}{\bf 5/16} & \multicolumn{5}{c||}{\bf 7/16} & \multicolumn{5}{c|}{\bf 9/16} \\
\hline\hline
1 & -- & \multicolumn{2}{c|}{--} & 87.07\,\% &  &  &  & \ding{108} &  & \ding{108} & \ding{108} & \ding{108} & \ding{108} & \ding{108} & \ding{108} & \ding{108} & \ding{108} & \ding{108} & \ding{108} & \ding{108} & \ding{108} & \ding{108} & \ding{108} & \ding{108} \\
2 & P & \multicolumn{2}{c|}{--} & 89.26\,\% &  &  & \ding{108} &  &  & \ding{108} &  & \ding{108} &  &  & \ding{108} & \ding{108} & \ding{108} &  & \ding{108} & \ding{108} & \ding{108} & \ding{108} & \ding{108} & \ding{108} \\
\cline{2-2}
3 & \multirow{14}{*}{\rotatebox{90}{P+O+S}} & \multicolumn{2}{c|}{--} & 92.27\,\% &  &  &  &  &  &  &  & \ding{108} &  &  &  & \ding{108} &  &  &  & \ding{108} & \ding{108} &  &  & \ding{108} \\
4 & & \multicolumn{2}{c|}{Crop(10\,\%)} & 92.31\,\% &  &  &  & \ding{108} &  &  &  &  &  &  &  &  &  &  &  &  & \ding{108} &  & \ding{108} &  \\
5 & & \multicolumn{2}{c|}{Crop(20\,\%)} & 91.58\,\% &  &  &  &  & \ding{108} &  &  &  & \ding{108} &  &  &  &  & \ding{108} &  & \ding{108} &  & \ding{108} & \ding{108} & \ding{108} \\
6 & & \multicolumn{2}{c|}{Noise(2.5\,\%)} & {\bf 92.40\,\%} &  & \ding{108} &  & \ding{108} &  &  & \ding{108} &  & \ding{108} &  &  &  & \ding{108} & \ding{108} & \ding{108} & \ding{108} &  & \ding{108} & \ding{108} & \ding{108} \\
7 & & \multicolumn{2}{c|}{Noise(5\,\%)} & 92.18\,\% & \ding{108} &  &  &  &  & \ding{108} & \ding{108} & \ding{108} &  & \ding{108} & \ding{108} & \ding{108} &  & \ding{108} & \ding{108} &  &  & \ding{108} &  & \ding{108} \\
8 & & \multicolumn{2}{c|}{Noise(10\,\%)} & 92.14\,\% &  &  &  &  &  &  &  &  &  &  &  &  &  &  &  &  & \ding{108} & \ding{108} &  &  \\
9 & & \multicolumn{2}{c|}{BodyModel(12)} & 91.71\,\% &  &  &  &  &  &  &  &  &  &  &  &  &  &  &  &  &  &  & \ding{108} &  \\
10 & & \multicolumn{2}{c|}{BodyModel(14)} & 91.92\,\% & \ding{108} & \ding{108} & \ding{108} &  & \ding{108} & \ding{108} & \ding{108} &  & \ding{108} & \ding{108} & \ding{108} & \ding{108} & \ding{108} & \ding{108} & \ding{108} & \ding{108} & \ding{108} & \ding{108} & \ding{108} & \ding{108} \\
11 & & \multicolumn{2}{c|}{KeyPose(10.6)} & 88.49\,\% &  &  &  &  &  &  &  &  &  &  & \ding{108} & \ding{108} & \ding{108} & \ding{108} & \ding{108} & \ding{108} & \ding{108} & \ding{108} &  &  \\
12 & & \multicolumn{2}{c|}{KeyPose(3.7)} & 91.67\,\% & \ding{108} & \ding{108} & \ding{108} &  & \ding{108} & \ding{108} & \ding{108} & \ding{108} & \ding{108} & \ding{108} & \ding{108} & \ding{108} & \ding{108} & \ding{108} & \ding{108} & \ding{108} & \ding{108} & \ding{108} & \ding{108} & \ding{108} \\
13 & & -- & Crop(10\,\%) & 90.85\,\% &  &  &  &  &  &  &  &  &  & \ding{108} & \ding{108} &  & \ding{108} &  &  & \ding{108} & \ding{108} &  &  &  \\
14 & & -- & Crop(20\,\%) & 85.78\,\% &  &  &  &  &  &  &  &  &  &  &  &  &  &  &  &  &  &  &  &  \\
15 & & Crop(10\,\%) & -- & 89.18\,\% &  &  &  &  &  &  &  &  &  &  &  &  &  &  &  &  &  &  & \ding{108} & \ding{108} \\
16 & & Crop(20\,\%) & -- & 85.83\,\% &  &  &  &  &  &  &  &  &  &  &  &  &  &  &  &  &  &  &  &  \\
\hline\hline
\multicolumn{5}{r|}{\rotatebox{90}{\bf Accuracy}\rotatebox{90}{\bf of fusion}} & \rotatebox{90}{93.51\,\%} & \rotatebox{90}{93.51\,\%} & \rotatebox{90}{93.47\,\%} & \rotatebox{90}{93.47\,\%} & \rotatebox{90}{93.43\,\%} & \rotatebox{90}{93.86\,\%} & \rotatebox{90}{93.81\,\%} & \rotatebox{90}{93.77\,\%} & \rotatebox{90}{93.73\,\%} & \rotatebox{90}{93.73\,\%} & \rotatebox{90}{\bf 94.03\,\%} & \rotatebox{90}{93.90\,\%} & \rotatebox{90}{93.86\,\%} & \rotatebox{90}{93.81\,\%} & \rotatebox{90}{93.81\,\%} & \rotatebox{90}{93.86\,\%} & \rotatebox{90}{93.77\,\%} & \rotatebox{90}{93.73\,\%} & \rotatebox{90}{93.73\,\%} & \rotatebox{90}{93.73\,\%} \\
\cline{6-25}
%
\end{tabular}
\end{table*}

\subsection{Retrospective Learning of All-in-One Model}

As soon as we efficiently identify the best combination of pre-processing techniques, a new question arises: Is it better to use the fusion of the independent classifiers, or to rather train a new \emph{all-in-one} classifier by all the normalized/augmented variants of actions obtained by the best identified techniques?

It is important to realize that the best combination can involve augmentation techniques that change the data format of original poses. In our case, the majority of best-performing combinations involves the BodyModel(14) technique that modifies the pose data format by decreasing the number of skeleton joints (from original 31 to 14). And it is hardly possible to mix different pose formats when training a single model. Consequently, we select the third 5/16 combination that reaches the 93.77\,\% fusion accuracy and involves the 5 techniques generating the same format of training/test actions: non-, P-, and P+O+S-normalized data along with P+O+S-normalized Noise(5\,\%) and KeyPose(3.7) augmentations.
We pre-process the actions by the given 5 techniques and use \emph{all} such 5 variants of training data to train a single all-in-one model. Then, we generate the same 5 variants of test data and evaluate the recognition power of the all-in-one model in two different ways: (i) each out of the 5 test-data variants is independently evaluated, which yields 5 recognition accuracies, or (ii) the proposed fusion approach is applied with the difference that individual variants of each test action are classified by the same all-in-one model.

The results of the first way of evaluation are reported in Table~\ref{tab:retroLearning}, where the best KeyPose(3.7) variant of test data reaches the accuracy of 90.64\,\%. This is significantly worse than the accuracy of 93.77\,\% achieved by the original fusion of independent classifiers. Regarding the second way of evaluation, each variant of a given test action is classified by the same all-in-one model to get the 5 partial classification outputs. Such outputs are then fused by the strict majority vote rule to get the test-action classification. In this case, we achieve the accuracy of 91.24\,\%, which is higher than the first way of evaluation (90.64\,\%), but still lower than the fusion approach with independent classifiers (93.77\,\%). This result allows us to answer the introductory question: It is better to fuse the results of independent models, rather than of a single all-in-one model, with respect to the same pre-processed variants of training/test data.
The reason is that the useful pre-processing techniques provide orthogonal views on the input data, which one neural-network model can hardly learn.



\begin{table}
\setlength{\tabcolsep}{4px}
\caption{Accuracy of 5 variants of test data evaluated on a single all-in-one Bi-LSTM model, which is trained on all the 5 variants of training data obtained by the same normalization/augmentation techniques.}
\label{tab:retroLearning}
\centering
\begin{tabular}{|c|c|r|}
\hline
\multicolumn{2}{|c|}{\bf Test data} & \multirow{2}{*}{\bf Accuracy} \\
{\bf Norm.} & {\bf Augmentation} & \\
\hline\hline
-- & -- & 89.52\,\% \\
P & -- & 89.13\,\% \\
\cline{1-1}
\multirow{3}{*}{\rotatebox{90}{P+O+S}} & -- & 90.46\,\% \\
& Noise(5\,\%) & 90.42\,\% \\
& KeyPose(3.7) & 90.64\,\% \\
\hline
%
\end{tabular}
\end{table}



\subsection{State-of-the-Art Comparison}

We have achieved the best result of 94.03\,\% in the fusion approach on the ground truth with 2,328 actions. Since the state-of-the-art results are reported on the ground truth with 2,345 actions, we simply consider that the rest of 17 actions are classified falsely, resulting in the accuracy of 93.35\,\%.
This accuracy is currently the clearly-best action recognition result reported on the HDM05 dataset -- see Table~\ref{tab:actionRecognitionStateOfTheArt}.
\begin{table}
\tabcolsep 4px
\centering
\caption{Comparison of the action-recognition accuracy of the state-of-the-art methods using the 2-fold cross validation (i.e., 50\,\% of training data) on the HDM05 dataset. The methods are sorted by the achieved accuracy.}
\label{tab:actionRecognitionStateOfTheArt}
\begin{tabular}{|p{7.5cm}||r|r|}
\hline
{\bf Method} & {\bf Acc.} \\
\hline\hline
%
LieNet-2Blocks~\cite{HWPG17} & 75.78 \\
CNN~\cite{LBTD17} & 83.33 \\
DMT-Net~\cite{ZZCZLZY20} & 85.30 \\
Si-GCN~\cite{LXZZCY19} & 85.45 \\
PGCN-TCA~\cite{YGZHZ20} & 86.59 \\
CNN features + $1$NN~\cite{SEZ18-mtap} & 86.79 \\
PB-GCN~\cite{TN18} & 88.17 \\
CNN feat. + $k$NN~\cite{SZ18-dexa} & 88.78 \\
%
Bi-LSTM + augm.~\cite{SZ19-ism} & 91.86 \\
\hline
{\bf Proposed approach (7/16 fusion)} & {\bf 93.35} \\
\hline
\end{tabular}
\end{table}


\section{Conclusions}

For the action recognition task, we have proposed the general approach for fusing independent classifiers and evaluating all their combinations efficiently. This independent-fusion approach has several advantages: (i) the majority vote rule enables selecting appropriate classifiers automatically for a given action, (ii) the data format of training actions can be different for individual classifiers in contrast to a single classifier, and (iii) the used Bi-LSTM architecture is completely independent, so it can be replaced by any kind of a neural network.
We believe that it is always better to train independent models for individual pre-processing techniques instead of a single all-in-one model, which can hardly learn orthogonal views on the input data.

By considering 16 variants of normalized/augmented input data, we have revealed that the combination of 7 Bi-LSTM classifiers clearly outperforms the state-of-the-art result on the challenging HDM05 dataset distinguishing the highest number of 130 classes, in comparison with other datasets.
%
%
Finally, we demonstrate the suitability of the proposed BodyModel and KeyPose augmentation techniques that are involved in the majority of the best-performing combinations of independent classifiers. This indicates that new augmentation techniques in combination with the fusion approach could increase the recognition accuracy of future classifiers.


\section*{Acknowledgments}
%
This research has been supported by the GACR project No. GA19-02033S.





\bibliographystyle{IEEEtran}
\bibliography{referencesdisa,references}




\end{document}